\def\year{2023}\relax
\pdfoutput=1
\documentclass[letterpaper]{article} 
\usepackage{aaai22}  
\usepackage{times}  
\usepackage{helvet}  
\usepackage{courier}  
\usepackage[hyphens]{url}  
\usepackage{graphicx} 
\usepackage{natbib}  
\usepackage{caption} 
\DeclareCaptionStyle{ruled}{labelfont=normalfont,labelsep=colon,strut=off} 
\usepackage[linesnumbered,titlenumbered,ruled,vlined,commentsnumbered,noend]{algorithm2e}

\usepackage{newfloat}
\usepackage{listings}
\usepackage[colorlinks,
linkcolor=red,
citecolor=blue
]{hyperref}

\usepackage{xcolor}
\usepackage{booktabs}
\usepackage{multirow}
\usepackage{threeparttable}
\usepackage{subfigure}
\usepackage{makecell}
\usepackage{pifont}
\usepackage[accsupp]{axessibility}  
\usepackage[export]{adjustbox}
\usepackage{amsfonts} 
\usepackage[rightcaption]{sidecap}

\newcommand{\figref}[1]{Fig.~\ref{#1}}
\newcommand{\reqref}[1]{Eq.~\eqref{#1}}
\newcommand{\secref}[1]{Sec.~\ref{#1}}
\newcommand{\tableref}[1]{Table~\ref{#1}}
\usepackage{xspace}
\makeatletter
\DeclareRobustCommand\onedot{\futurelet\@let@token\@onedot}
\def\@onedot{\ifx\@let@token.\else.\null\fi\xspace}
\def\eg{\emph{e.g}\onedot} 
\def\ie{\emph{i.e}\onedot} 
 
\def\etc{\emph{etc}\onedot} 
 
\def\etal{\emph{et al}\onedot}
\makeatother

\definecolor{americanrose}{rgb}{1.0, 0.01, 0.24}

\title{Background-Mixed Augmentation for Weakly Supervised Change Detection}

\author{
    Rui Huang\textsuperscript{\rm 1}, Ruofei Wang\textsuperscript{\rm 1}, Qing Guo\textsuperscript{\rm 2,3} \thanks{Qing Guo is the corresponding author: {tsingqguo@ieee.org}.},
    Jieda Wei\textsuperscript{\rm 1},
    Yuxiang Zhang\textsuperscript{\rm 1}, Wei Fan\textsuperscript{\rm 1},
    Yang Liu\textsuperscript{\rm 4,5}
}
\affiliations{
    \textsuperscript{\rm 1} School of Computer Science and Technology, Civil Aviation University of China, China\\
    \textsuperscript{\rm 2} IHPC, Agency for Science, Technology and Research, Singapore \\
    \textsuperscript{\rm 3} CFAR, Agency for Science, Technology and Research, Singapore \\
    \textsuperscript{\rm 4} Zhejiang Sci-Tech University, China\\
    \textsuperscript{\rm 5} Nanyang Technological University, Singapore\\
    \{rhuang, yxzhang, wfan\}@cauc.edu.cn, \{rf\_wang, jdwei1\}@yeah.net, tsingqguo@ieee.org, yangliu@ntu.edu.sg

}

\usepackage{bibentry}

\begin{document}

\maketitle

\begin{abstract}
Change detection (CD) is to decouple object changes (\ie, object missing or appearing) from background changes (\ie, environment variations) like light and season variations in two images captured in the same scene over a long time span, presenting critical applications in disaster management, urban development, \etc.
In particular, the endless patterns of background changes require detectors to have a high generalization against unseen environment variations, making this task significantly challenging.
Recent deep learning-based methods develop novel network architectures or optimization strategies with paired-training examples, which do not handle the generalization issue explicitly and require huge manual pixel-level annotation efforts.
In this work, for the first attempt in the CD community, we study the generalization issue of CD from the perspective of data augmentation and develop a novel weakly supervised training algorithm that only needs image-level labels.
Different from general augmentation techniques for classification, we propose the \textit{background-mixed augmentation} that is specifically designed for change detection by augmenting examples under the guidance of a set of background changing images and letting deep CD models see diverse environment variations.
Moreover, we propose the \textit{augmented~\&~real data consistency loss} that encourages the generalization increase significantly.
Our method as a general framework can enhance a wide range of existing deep learning-based detectors.
We conduct extensive experiments in two public datasets and enhance four state-of-the-art methods, demonstrating the advantages of our method.
We release the code at \url{https://github.com/tsingqguo/bgmix}.
\end{abstract}

\section{Introduction}
\label{sec:intro}

\begin{figure}[!tb]
    \centering
    \includegraphics[width=\linewidth]{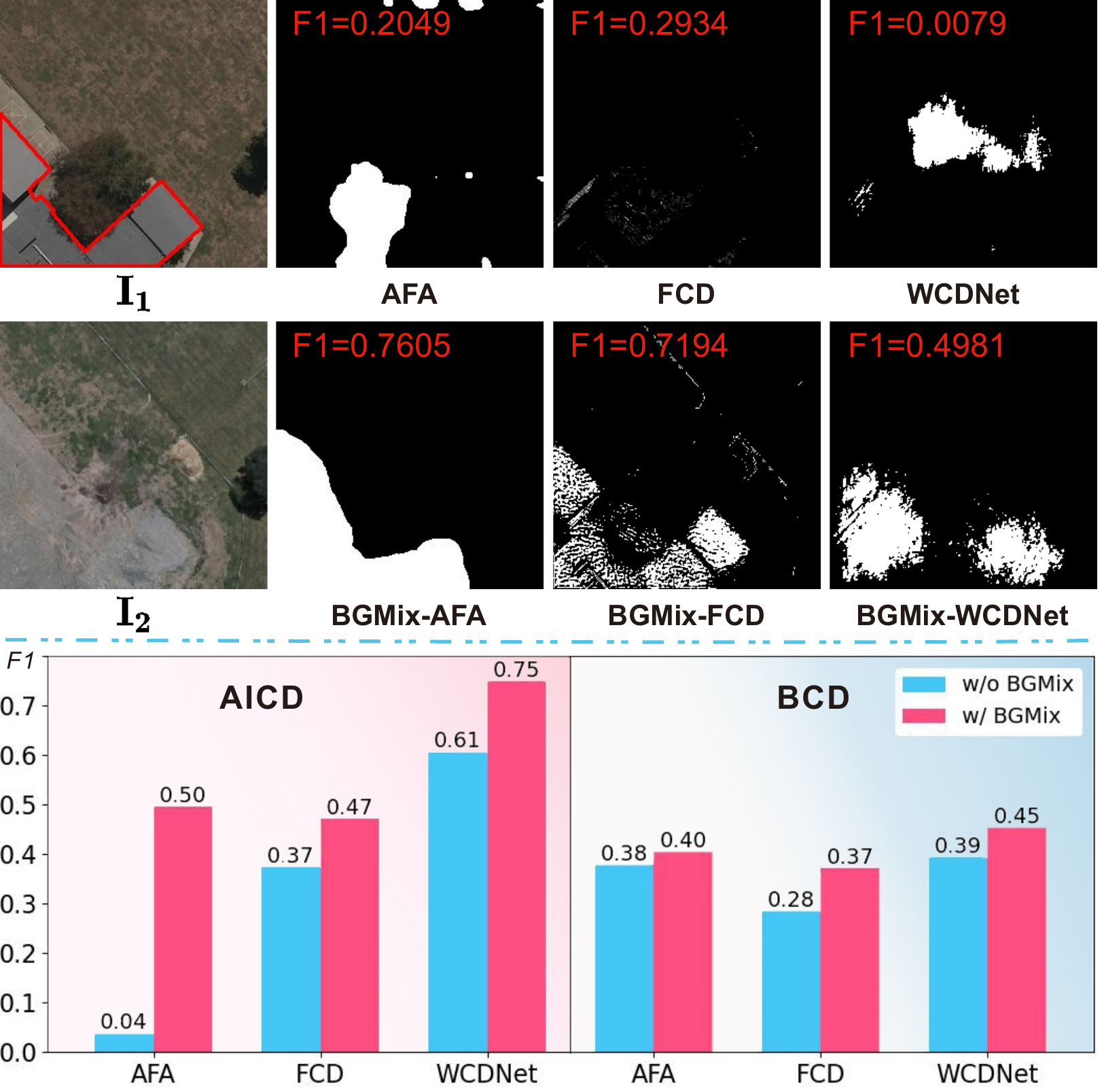}
\caption{Comparison of three SOTA methods with/without our BGMix. Clearly, BGMix improves the three methods with significant F1 increases on both AICD dataset~\cite{bourdis2011constrained} and BCD dataset~\cite{ji2018fully}, presenting more accurate segmentation results. }
    \label{fig:my_mot}
\end{figure}

Change detection (CD) is to segment object changes (\ie, object missing or appearing) from background changes (\ie, environment variations) like light and season variations in two images captured in the same scene with a long time span (See examples shown in \figref{fig:my_mot}), presenting critical applications in the disaster management~\cite{sublime2019automatic},  urban development~\cite{lee2021local}, resource monitoring, and utilization~\cite{ye2021near}, and visual surveillance~\cite{goyette2012changedetection}, \etc.

Change detection is still challenging and far from real-world applications due to the complex and unknown environment variations.
For example, in \figref{fig:my_mot}, there exist various differences on the lawn (\eg, color and density differences) between two images, which should be identified as background changes and not be segmented. Nevertheless, the state-of-the-art (SOTA) method WCDNet \cite{andermatt2020weakly} focuses on the lawn variations while missing the desired object changes (\ie, the disappearance of a building).
Other environment variations like season and illumination variations cause similar issues.
The endless patterns of background changes require detectors to have a high generalization against unseen environment variations.

In recent years, a series of network architectures are designed with convolutional \cite{zhang2021object,zheng2021building} and transformer layers \cite{chen2021remote,bandara2022transformer} to extract effective features to represent the changes.
Although achieving impressive progress, they do not handle the generalization issue explicitly and usually require huge pixel-level annotation efforts.
The weakly supervised strategy is an effective solution to alleviate the pixel-level annotation efforts \cite{andermatt2020weakly,wu2022fully,ru2022learning}.
However, without pixel-level supervision, change detectors suffer from low detection accuracy due to complex background changes. 
%
%
The root reason for this dilemma is the training data does not cover diverse environment variations. 


In this work, for the first attempt in the CD community, we study the generalization issue from the perspective of data augmentation and develop a novel weakly supervised algorithm that only needs image-level labels.
Different from general augmentation methods for classification, we propose the \textit{background-mixed augmentation} that is specifically designed for change detection by augmenting examples under the guidance of a set of background images and letting targeted deep models see diverse background changes.
Moreover, we propose the \textit{augmented~\&~real data consistency loss} that encourages the high generalization significantly.
Our method as a general framework can enhance a wide range of existing deep learning-based detectors. 
As presented in Fig.~\ref{fig:my_mot}, with our augmentation method, the performance of three change detectors is improved with large margins.
Our contributions can be summarized as follows:
\begin{itemize}
    \item We study the generalization issue of the CD from the data augmentation perspective and propose a novel weakly supervised change detection method denoted as BGMix.
    \item We propose a \textit{background-mixed augmentation} that is costumed for change detection and \textit{augmented~\&~real data consistency loss} to encourage the generalization. 
    \item We conduct extensive experiments in two public datasets and enhance four state-of-the-art methods, demonstrating the advantages of our method.
\end{itemize}

\section{Related Work}
\label{sec:related}

\textbf{Change detection methods.}
Conventional change detection methods are based on simple algebraic methods, such as gradient~\cite{di2003change}, change vector analysis~(CVA)~\cite{malila1980change} 
and thresholding~\cite{rosin1998thresholding}. A simple strategy cannot deal with background change noises. More complex models are introduced such as Markovian data fusion~\cite{moser2011multiscale} and dictionary learning~\cite{gong2016coupled}. To obtain a high-quality CD, \cite{feng2015fine} and \cite{stent2016precise} also propose image alignment and lighting correction before CD.
With the development of deep learning, different deep models are introduced into CD, such as restricted Boltzmann machine~(RBM)~\cite{gong2016change}, CNN~\cite{ding2016automatic}, GAN~\cite{kousuke2017use,gong2017generative}, RCNN~\cite{mou2019learning}, LSTM~\cite{lyu2016learning}, and Transformer~\cite{chen2021remote}. Compared with RBM and GAN, it is easier to design variant CNN-based network architectures for CD. A representative network, ADCDnet ~\cite{huang2020change} proposes to use absolute difference features of multiscale convolutional layers for CD. To extract more effective features, position- and channel-correlated attentions are incorporated into CD by Zhang \etal~\cite{zhang2021object}. The current trend is using transformers for CDs. ChangeFormer~\cite{bandara2022transformer} unifies hierarchically structured transformer encoder with Multi-Layer Perception (MLP) decoder in Siamese network architecture to render multi-scale long-range details for CD. The success of the above deep change detectors is due in large part to a huge number of pixel-level labeled images. However, the existing CD datasets cannot contain all situations of environment changes, which limits the generalization ability of the recent change detectors.

\textbf{Weakly supervised detection methods.}
The weakly supervised method is an effective way to conquer the needs of pixel-level data. Weakly Supervised Change Detection (WSCD) aims to detect pixel-level changes by using the image-level labels ~\cite{ijcai2017p279,krahenbuhl2011efficient}.
Sakurada \etal~\cite{sakurada2020weakly} divide the semantic change detection to change detection and semantic segmentation and do not need the semantic change annotations.
%
Andermatt \etal~\cite{andermatt2020weakly} design a custom remapping block and employ a CRF-RNN to refine the change mask.
Kalita \etal~\cite{kalita2021land} combine the PCA and K-means algorithm to predict changes from the image difference generated by a Siamese network.
With the assumption that changed image pairs can be discriminated as unchanged image pairs by masking out change regions, 
Wu \etal~\cite{wu2022fully} propose a generative adversarial network (GAN) to detect changes.
Similar to fully supervised CD, WSCD also suffers from the low generalization issue.
The root reason for this dilemma is that the training data cannot contain various environmental changes.

\begin{figure*}[!htb]
    \centering
    \includegraphics[width=\linewidth]{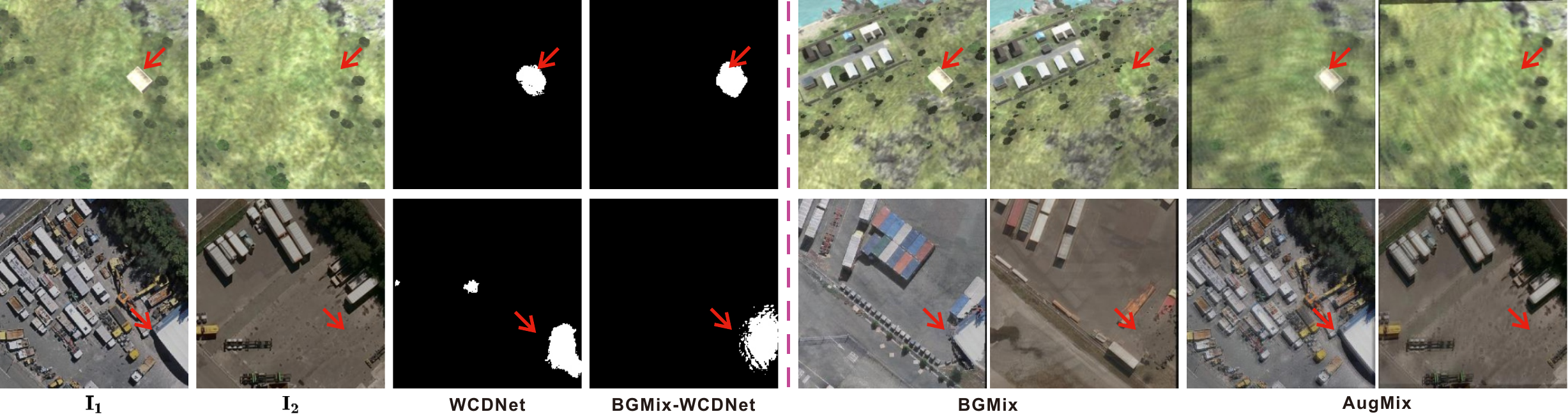}
    \caption{Comparison of augmented examples between the AugMix and BGMix. AugMix will transform the whole image pixels randomly, while BGMix can effectively mix the background noise (\ie, cars, buildings) with change regions. We use a state-of-the-art method (\ie, WCDNet~\cite{andermatt2020weakly}) to study the effect of BGMix, and change regions are pointed by the red arrows.}
    \label{fig:limitations}
\end{figure*}

\textbf{Data augmentation techniques.}
Data augmentation is an effective solution to improve the generalization capability of deep models.
Cutout \cite{devries2017improved} randomly masks out square regions of images to regularize the network training.
Random erasing \cite{zhong2020random} replaces the pixels with random intensities of a randomly selected rectangle region.
Cutout or random erasing might result in information loss. Then, 
CutMix~\cite{yun2019cutmix} cuts patches and pastes them among the training images.
Later works pay more attention to mixing, such as MixUp \cite{huang2020self}, AugMix \cite{hendrycks2020augmix}, SnapMix \cite{huang2021snapmix}, AlignMixup \cite{venkataramanan2021alignmixup} \etc.
Compared with random cropping, SuperMix \cite{dabouei2021supermix} and KeepAugment \cite{gong2021keepaugment} preserve the informative regions~(\ie salient regions). 
AutoAugment \cite{cubuk2018autoaugment} automatically searches augmentation policies in a predefined search space.
Deeprepair \cite{yu2021deeprepair} employs the style transformation to use some failure examples for augmentation guidance.
There are also augmentation methods for other tasks like deraining \cite{guo2020efficientderain} and night2day transformation \cite{fu2021let}. 
Existing data augmentation techniques are not suitable for change detection.
CD is to decouple object changes from environmental changes. A good data augmentation method for CD should not only simulate the object changes but also simulate various environment changes.

\section{Problem Formulation and Motivation}
\label{sec:problemformulation}


Given two images $\mathbf{I}_1$ and $\mathbf{I}_2$ that are captured in the same scene with a long time span, the change detection task is to identify object changes in the two images. For example, when an object in $\mathbf{I}_1$ disappears in $\mathbf{I}_2$, a CD method is desired to output a binary map (\ie, $\mathbf{C}$) where the missed object region is identified as one and other positions are assigned zero. Recent deep learning-based methods formulate this task as an image2image mapping task via a convolution neural network (CNN) \cite{lei2020hierarchical}, \eg, 
%
\begin{align} \label{eq:cd_cnn}
\mathbf{C} = \phi_\theta(\mathbf{I}_1, \mathbf{I}_2)
\end{align}
%
where $\phi_\theta(\cdot)$ is the CNN and $\theta$ denotes its parameters. The tensor $\mathbf{C}$ denotes the predicted changing mask. A straightforward implementation is to train the network via paired training examples, \ie, $\mathcal{T}_\text{train}=\{<\mathbf{I}_1, \mathbf{I}_2,\mathbf{C}_\text{gt}>\}$ via loss functions like binary cross-entropy, mean squared error~(MSE), or $L_1$, where $\mathbf{C}_\text{gt}$ denotes the ground truth of the example.
However, there are two limitations preventing such a solution for real-world applications: \ding{182} it requires laborious human efforts to label the pixel-wise ground truth, which can hardly construct large-scale datasets and the potential labeling noise might affect the model training. \ding{183} the training strategy does not consider the generalization to unseen environment changes (\ie, background variance) that have diverse patterns in different images.

An alternative learning method is the weak supervision based on the generative adversarial network (GAN) with image-level labels \cite{wu2022fully} that only indicates whether an image pair contains object changes or not. 
Specifically, we use a CD model to predict the changing mask of an image pair, and then crop the changing regions from the image pair. 
After that, we can feed the cropped image pair to a discriminator that estimates whether object changes remain.
Intuitively, when the changes are previously identified by the CD model and removed, the cropped image pair would not contain any object changes.
%


Although effective, existing weakly supervised change detection methods neglect the potential environment changes within a long time span, which might be caused by light variation, season variation, weather variation, \etc. As a result, they easily generate obvious detection errors on unseen background variations. We take the recent weakly supervised change detection method as an example and use it to handle different image pairs. As shown in \figref{fig:limitations}, the method can get accurate detection results when the background changes are minor (See the first case in \figref{fig:limitations}) but produce obvious errors when the background changes have obvious color shifting (\eg, the second case.) Existing data augmentations (\eg, AugMix~\cite{hendrycks2020augmix}) for image classification change the whole images at 
a time and present less effectiveness on background enrichment.

\begin{SCfigure*}
    \centering
    \includegraphics[width=1.65\linewidth]{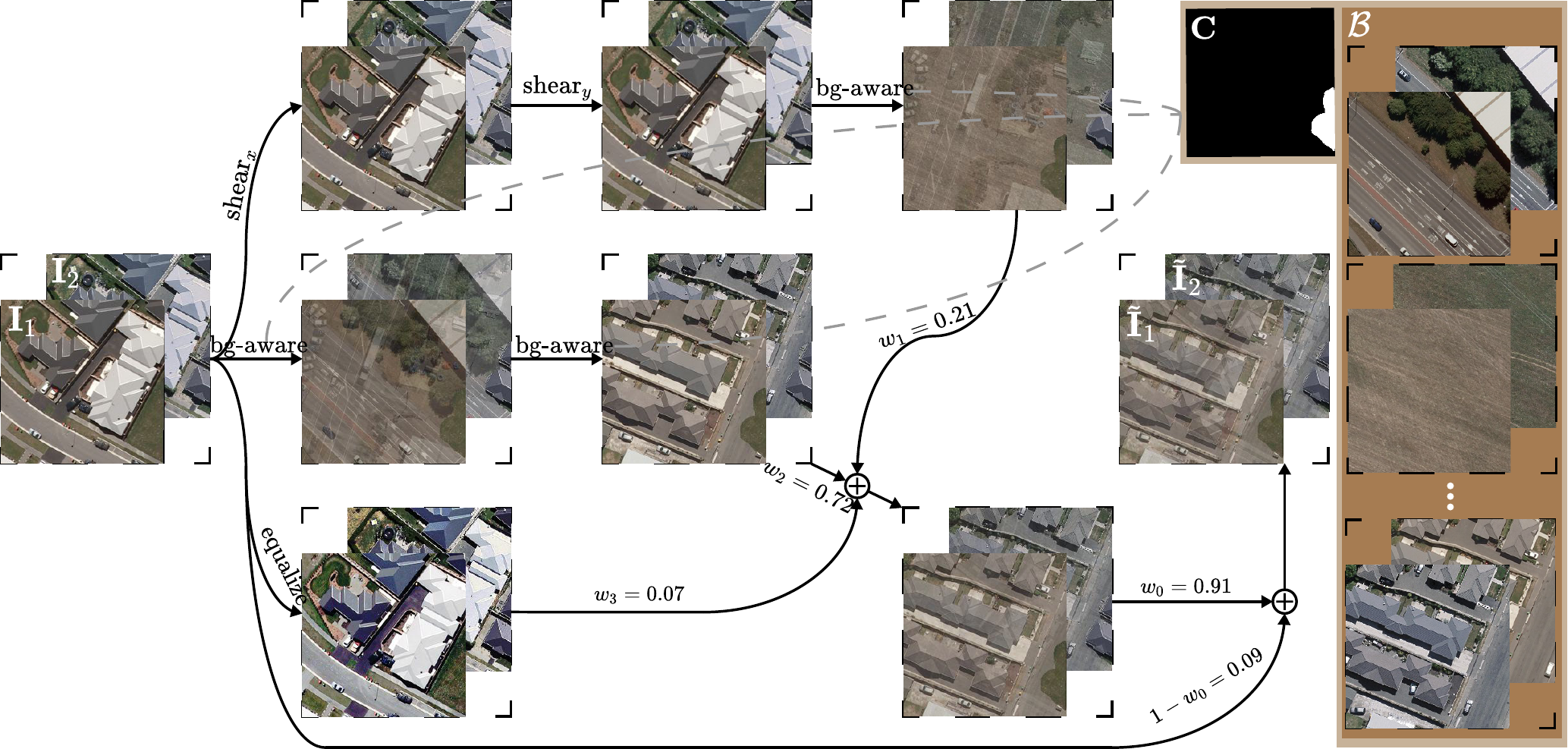}
    \caption{Pipeline of the background-mixed augmentation introduced in \secref{subsec:backaug}. `bg-aware' denotes the background-aware augmentation operation (See \secref{subsec:backaug}) taking the background pair and the changing mask as inputs.
    }
    \label{fig:framework}
\end{SCfigure*}

\section{Methodology}
\label{sec: ba-aug}

\subsection{Overview}

To address the limitations, we propose a background-mixed augmentation (BGMix) to train a deep model (\ie, $\phi_\theta(\cdot)$) for change detection in a weakly supervised way. 
Before going to the main process, we first detail the required inputs:
\begin{itemize}
     \item {\bf Weakly supervised training set $\mathcal{T}_\text{train}$.} We require a training dataset (\ie, $\mathcal{T}_\text{train}=\{<\mathbf{I}_1, \mathbf{I}_2>\}^T$) with $T$ examples and image-level labels. We name it a weakly supervised training dataset since we only know whether there exist object changes between $\mathbf{I}_1$ and $\mathbf{I}_2$ and do not require the pixel-wise annotations.
    \item {\bf Background-guided set $\mathcal{B}$.} We collect $N$ image pairs where each image pair (\eg, $<\mathbf{B}_1,\mathbf{B}_2>$) only contains background changes without any object changes. 
    We denote the set as background-guided set, \ie, $\mathcal{B}=\{<\mathbf{B}_1,\mathbf{B}_2>\}^N$ (See \figref{fig:framework} for three background changing images as examples). These background pairs have distinct appearances and are used to enrich the potential background changing patterns in the training examples.
    \item {\bf Augmentation operation set $\mathcal{O}$.} We set an operation set $\mathcal{O}=\{\text{o}_1,\ldots,\text{o}_M\}$ that contains $M$ augmentation operations. Each operation is used to transform the image pair $<\mathbf{I}_1,\mathbf{I}_2>$ to a novel one (\ie, $<\mathbf{I}_1^{\text{o}_j},\mathbf{I}_2^{\text{o}_j}>$), \eg,
    %
    \begin{align} \label{eq:bg_sys}
    \mathbf{I}_1^{\text{o}_j},\mathbf{I}_2^{\text{o}_j} = \text{o}_j(\mathbf{I}_1, \mathbf{I}_2,\alpha_j),
    \end{align}
    %
    where $\alpha_j$ denotes the operation-related parameters. For example, if we set the $j$th $\text{o}_j(\cdot)$ as the translation operation, $\alpha_j$ means the shifting distance and direction of the two images and the output images (\ie, $\mathbf{I}_1^{\text{o}_j}$ and $\mathbf{I}_2^{\text{o}_j}$) are the shifted counterparts of the input images. Beyond the simple geometry augmentations like translation, we propose a specific operation (\ie, \textit{background-aware augmentation operation}) in \secref{subsec:backaug} for the CD task, which helps the CD models see different background changes.
\end{itemize}
Intuitively, BGMix aims to utilize the background-guided set $\mathcal{B}$, the predicted changing mask $\mathbf{C}$, and the image pair with $\mathbf{I}_1$ and $\mathbf{I}_2$ to generate new image pairs with \textit{diverse} background patterns. 
As a result, the targeted CNN can see different background interference, and then its generalization could be enhanced. 
%

%
During the training process, we have a model $\phi_\theta$ updated at the previous iteration and sample a training example (\eg, $<\mathbf{I}_1, \mathbf{I}_2>$) from $\mathcal{T}_\text{train}$.
Then, we can extract a changing mask by $\mathbf{C}=\phi_\theta(\mathbf{I}_1, \mathbf{I}_2)$.
After that, we use the proposed augmentation strategy to generate an augmented image pair $<\tilde{\mathbf{I}}_1, \tilde{\mathbf{I}}_2>$ as detailed in \secref{subsec:backaug} with the pipeline shown in \figref{fig:framework}.
%
Finally, we propose novel loss functions (\ie, augmented~\&~real data consistency loss) in \secref{subsec:augloss} to encourage the CNN to predict similar results across the original image pairs and background-augmented image pairs.
%
We also detail the whole process in \secref{subsec:algorithm} with Algorithm~\ref{alg}.

\subsection{Background-mixed Augmentation}
\label{subsec:backaug}
%

\textbf{Background-aware augmentation operation.} We augment a training example (eg, $<\mathbf{I}_1, \mathbf{I}_2>$) by enriching the background changes between the image pair. 
Instead of the common augmentations like translation or zooming, we use the collected background-guided set (\ie, $\mathcal{B}$) to replace the original background of the input image pair according to the changing mask $\mathbf{C}$. 
Specifically, we first sample a background pair (\ie, $<\mathbf{B}_1,\mathbf{B}_2>$) from the background-guided dataset (\ie, $\mathcal{B}$) randomly, and then we transform the input image pair with
%
\begin{align} \label{eq:bg_aug_oveview}
   \mathbf{I}_1^{\text{o}_j},\mathbf{I}_2^{\text{o}_j} = \text{o}_j(\mathbf{I}_1, \mathbf{I}_2,\alpha_j=\{\mathbf{B}_1,\mathbf{B}_2,\mathbf{C}\}),
\end{align}
%
which can be detailed as
%
\begin{align} 
    \label{eq:bg_aug1}
    \mathbf{I}_1^{\text{o}_j} & = \text{Rep}(\mathbf{I}_1, \mathbf{B}_1, \mathbf{C}) = \mathbf{I}_1 \odot \mathbf{C} + \mathbf{B}_1 \odot (1-\mathbf{C}), \\
     \label{eq:bg_aug2}
    \mathbf{I}_2^{\text{o}_j} & = \text{Rep}(\mathbf{I}_2, \mathbf{B}_2, \mathbf{C}) = \mathbf{I}_2 \odot \mathbf{C} + \mathbf{B}_2 \odot (1-\mathbf{C}),
\end{align}
%
where `$\odot$' is the element-wise multiplication. 
The function $\text{Rep}(\mathbf{I}_1, \mathbf{B}_1, \mathbf{C})$ is to replace the region in $\mathbf{I}_1$ indicated by $(1-\mathbf{C})$ with the corresponding region in $\mathbf{B}_1$.
Note that, compared to common augmentation operations, the above method changes the background region of the image pair while maintaining the object changing region, leading to new image pairs with different background changes.

\textbf{Augmentation strategy (\ie, Line 1-11 in Algorithm \ref{alg}).}  Inspired by recent augmentation methods for classification \cite{hendrycks2020augmix}, we conduct multi-depth random augmentations on the image pair to guarantee the diversity of background variations. Specifically, we set $K$ augmentation paths. For each patch, we first sample three operations for $\mathcal{O}$ (See Line 5 in Algorithm \ref{alg}) and stack them to construct new operations (See Line 6 in Algorithm \ref{alg}). Then, we randomly sample one operation from the newly constructed operations, which is used to augment the input image pair $\mathbf{I}_1, \mathbf{I}_2$ with the $\mathcal{B}$ and change mask $\mathbf{C}$ and get a new image pair. To conduct $K$ times augmentation, we get $K$ new image pairs that are mixed with randomly sampled weights (See Line 3 in Algorithm \ref{alg}). Finally, the mixed image pair is further blended with the original input (See Line 9 in Algorithm \ref{alg}) and get the final output, \ie, $<\tilde{\mathbf{I}}_1,\tilde{\mathbf{I}}_2>$.

\begin{algorithm}[tb]
	{
		\caption{\small{Learning CD models via BGMix}}\label{alg}
		\KwIn{$\phi_\theta(\cdot)$, Loss function $\mathcal{L}_\text{con}$, Training dataset $\mathcal{T}_\text{train}=\{<\mathbf{I}_1, \mathbf{I}_2>\}$, Background-guided set $\mathcal{B}$, Operation set $\mathcal{O}$=\{background-aware operation,rot, shear,trans, equalize\}, and maximum epoch number (\ie, MaxIters).}
		\KwOut{Trained CD model $\phi_\theta(\cdot)$.}
        \SetKwFunction{FMain}{BGMix}
        \SetKwProg{Fn}{Function}{:}{}
        \Fn{\FMain{$\mathcal{B},\mathbf{I}_1,\mathbf{I}_2,\mathbf{C}$}}{
        Initialize two empty maps $<\tilde{\mathbf{I}}_1,\tilde{\mathbf{I}}_2>$\;
        Sample mixing weights $(w_1,w_2,\ldots,w_K)\sim\text{Dirichlet}$\;
        \For{$i=1\ \mathrm{to}\ K$}{
            Sample operations $(\text{o}_1,\text{o}_2,\text{o}_3)\sim\mathcal{O}$\;
            Combine via $\text{o}_{12}=\text{o}_2\text{o}_1$ and $\text{o}_{123}=\text{o}_3\text{o}_2\text{o}_1$\;
            Sample $\text{o}\sim\{\text{o}_{1},\text{o}_{12},\text{o}_{123}\}$\;
            $\tilde{\mathbf{I}}_1+=w_i\text{o}(\mathbf{I}_1,\mathcal{B},\mathbf{C})$, $\tilde{\mathbf{I}}_2+=w_i\text{o}(\mathbf{I}_2,\mathcal{B},\mathbf{C})$ \;
            }
        Sample a blending weight $w_0\sim\text{Beta}$\;
        \textbf{return} $\{\tilde{\mathbf{I}}_j=w_0\mathbf{I}_j+(1-w_0)\tilde{\mathbf{I}}_j|j\in\{1,2\}\}$\;
        }
        \textbf{End function}\;
 		\For{$t=1\ \mathrm{to}\ \mathrm{MaxIters}$}{
 		    Sample an image pair via $<\mathbf{I}_1,\mathbf{I}_2>\sim \mathcal{T}_\text{train}$\;
 		    Perform $\mathbf{C}=\phi_\theta(\mathbf{I}_1,\mathbf{I}_2)$\;
 		    Generate an augmented image pair via $<\tilde{\mathbf{I}}_1,\tilde{\mathbf{I}}_2>=\FuncSty{BGMix}(\mathcal{B},\mathbf{I}_1,\mathbf{I}_2,\mathbf{C})$\;
 		    Sample a background pair $<\mathbf{B}_1,\mathbf{B}_2>\sim \mathcal{B}$\;
 		    Generate $<\mathbf{I}_1',\mathbf{I}_2'>$, $<\tilde{\mathbf{I}}_1',\tilde{\mathbf{I}}_2'>$, $<\mathbf{B}_1',\mathbf{B}_2'>$ \& $<\tilde{\mathbf{B}}_1',\tilde{\mathbf{B}}_2'>$ via \reqref{eq:bg_replace1}, and \eqref{eq:bg_replace3}\;
 		    Perform
 		    $\tilde{\mathbf{C}}=\phi_\theta(\tilde{\mathbf{I}}_1,\tilde{\mathbf{I}}_2)$,
 		    ${\mathbf{C}}'=\phi_\theta(\mathbf{B}_1,\mathbf{B}_2)$
 		    \;
            Calculate loss function via \reqref{eq:consist_loss} \;
            Do back-propagation and update $\phi_\theta$\;
    	}
	}
\end{algorithm}

\subsection{Augmented~\&~Real Data Consistency Loss}
\label{subsec:augloss}

Intuitively, the CNN for change detection is desired to predict the same results between augmented image pairs (\ie, $<\tilde{\mathbf{I}}_1,\tilde{\mathbf{I}}_2,l>$) with different backgrounds and the original image pairs (\ie, $<\mathbf{I}_1,\mathbf{I}_2,l>$). 
We propose the augmented~\&~real data consistency loss to encourage this property explicitly, which contains five parts.

\textit{First}, we want to encourage the perception similarity between~$<\mathbf{I}_1^i,\mathbf{I}_2^i>$ and $<\tilde{\mathbf{I}}_1^i,\tilde{\mathbf{I}}_2^i>$ as the first loss function, \ie,
%
\begin{align} \label{eq:consist_loss1}
\mathcal{L}_\text{con1} & =
\Phi(\mathbf{I}_1,\mathbf{I}_2,\tilde{\mathbf{I}}_1,\tilde{\mathbf{I}}_2) \nonumber \\
& = 1- \text{Cos}(\psi([\mathbf{I}_1,\mathbf{I}_2]),\psi([\tilde{\mathbf{I}}_1,\tilde{\mathbf{I}}_2]))
\end{align}
%
where `$[\cdot]$' is the concatenation operation, `$\psi(\cdot)$' means the pre-trained VGG16 network for the perception feature extraction, and `$\text{Cos}(\cdot)$' is the cosine similarity function. The hyper-parameter $\lambda_1$ is to weight the consistency loss.
The intuitive idea behind this loss is that the two image pairs $<\mathbf{I}_1,\mathbf{I}_2>$ and $<\tilde{\mathbf{I}}_1,\tilde{\mathbf{I}}_2>$ should contain the same object changes if the predicted mask $\mathbf{C}$ is accurate. 
For example, if $\mathbf{C}$ mislabels the change pixels between $\mathbf{I}_1$ and $\mathbf{I}_2$ as unchanged pixels, $[\mathbf{I}_1,\mathbf{I}_2]$ and $[\tilde{\mathbf{I}}_1,\tilde{\mathbf{I}}_2]$ would present distinct appearance on the foreground change regions since the latter image pair is produced according to the $\mathbf{C}$ (See \reqref{eq:bg_aug1}).

\textit{Second}, to further allow the CNN to be robust to different background changes, we randomly sample a background pair $<\mathbf{B}_1,\mathbf{B}_2>$ from the set $\mathcal{B}$. Then, we use it to replace the backgrounds of $<\mathbf{I}_1,\mathbf{I}_2>$ and get a new image pair by
\begin{align} 
\label{eq:bg_replace1}
\mathbf{I}_1' = \text{Rep}(\mathbf{I}_1,\mathbf{B}_1,\mathbf{C}), 
\mathbf{I}_2' = \text{Rep}(\mathbf{I}_2,\mathbf{B}_2,\mathbf{C}), 
\end{align}
%
where the function $\text{Rep}(\cdot)$ is the same with one defined in \reqref{eq:bg_aug1}.
Moreover, we can synthesize a new background pair by pasting the  background regions in the $<\mathbf{I}_1,\mathbf{I}_2>$ to the corresponding positions in $<\mathbf{B}_1,\mathbf{B}_2>$. 
Then, we can get a new background pair like \reqref{eq:bg_replace1}, \ie,
%
\begin{align} 
\label{eq:bg_replace3}
\mathbf{B}_1' = \text{Rep}(\mathbf{B}_1,\mathbf{I}_1,\mathbf{C}), 
\mathbf{B}_2' = \text{Rep}(\mathbf{B}_2,\mathbf{I}_2,\mathbf{C}). 
\end{align}
%
Intuitively, if $\mathbf{C}$ localizes object changes accurately, $<\mathbf{I}_1,\mathbf{I}_2>$ and $<\mathbf{I}_1',\mathbf{I}_2'>$ should present the same perceptions, similar on
$<\mathbf{B}_1,\mathbf{B}_2>$~\&~$<\mathbf{B}_1',\mathbf{B}_2'>$, and we have
%
\begin{align} \label{eq:consist_loss2}
\mathcal{L}_\text{con2} & = \Phi(\mathbf{I}_1,\mathbf{I}_2,\mathbf{I}_1',\mathbf{I}_2')
+ \Phi(\mathbf{B}_1,\mathbf{B}_2,\mathbf{B}_1',\mathbf{B}_2')
\end{align}
%
where $\Phi(\cdot)$ is the same as the function in \reqref{eq:consist_loss1}.

\textit{Third}, we encourage the background replacement while utilizing the low area rate prior of object changes, that is,
%
\begin{align} \label{eq:consist_loss3}
\mathcal{L}_\text{con3}  = \sum_{i=1}^2 (1-\text{SSIM}(\mathbf{I}_i, \mathbf{I}_i')-L_1(\mathbf{I}_i, \mathbf{I}_i')),
\end{align}
%
where $\text{SSIM}(\cdot)$ is the structure similarity loss.  Minimizing the above loss function leads to two results according to the \reqref{eq:bg_replace1}: \ding{182} More background contexts from the background pair are embedded and it benefits the augmentation. \ding{183} The object changes in an image pair are usually small and take a low area rate in the whole image (See \figref{fig:limitations}). The proposed loss function makes use of this prior to encouraging small change areas in $\mathbf{C}$.
%
%
\begin{SCfigure*}
    \centering
    \includegraphics[width=1.8\linewidth]{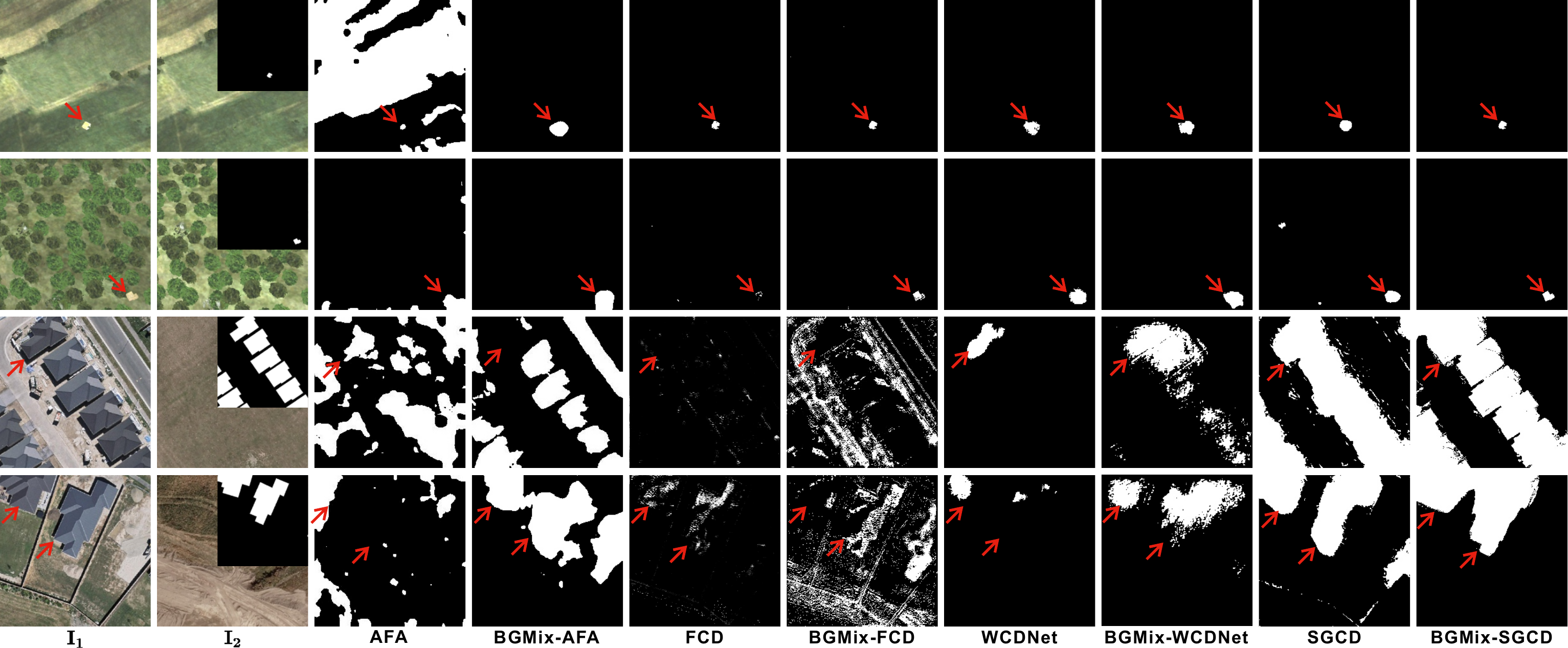}
    \caption{Visual comparison of CD results detected by different methods with and without BGMix. The change objects are marked out by the red arrows.}
    \label{fig:my_vision}
\end{SCfigure*}

\textit{Fourth}, we further employ two discriminators as adversarial losses to encourage the CNN $\phi_\theta(\cdot)$ to produce accurate change masks.
Intuitively, if the change mask $\mathbf{C}$ identifies the change regions accurately, the discriminators cannot distinguish $<\mathbf{I}_1',\mathbf{I}_2'>$ from $<\mathbf{I}_1,\mathbf{I}_2>$, and $<\mathbf{B}_1',\mathbf{B}_2'>$ from $<\mathbf{B}_1,\mathbf{B}_2>$, respectively. Specifically, the loss function could be represented as
\begin{align} \label{eq:consist_loss4}
    \mathcal{L}_\text{con4} &=
    %
    \text{log}(\text{D}_1(\mathbf{I}_1,\mathbf{I}_2))
        + \text{log}(1-\text{D}_1(\mathbf{I}_1',\mathbf{I}_2')) 
        \nonumber\\
        & + \text{log}(\text{D}_2(\mathbf{B}_1,\mathbf{B}_2))
        + \text{log}(1-\text{D}_2(\mathbf{B}_1',\mathbf{B}_2')),
\end{align}
where $\text{D}_1$ and $\text{D}_2$ are the two discriminators. It will be minimized when the function $\phi_\theta(\cdot)$ can detect the $\mathbf{C}$ accurately, which means the synthesized image pairs in \reqref{eq:bg_replace1} and \reqref{eq:bg_replace3} are close to realistic image pairs $<\mathbf{I}_1,\mathbf{I}_2>$ and $<\mathbf{B}_1,\mathbf{B}_2>$ that discriminators can not distinguish them.

\textit{Finally}, to suppress the potential predicting errors caused by background changes, we set a loss function to encourage $\phi_\theta(\cdot)$ to predict an all-zero mask when we take a background pair as inputs, \ie,
\begin{align}
    \mathcal{L}_\text{con5}= L_2(\phi_\theta(\mathbf{B_1},\mathbf{B_2}), \mathbf{0})
    \label{eq:consist_loss5}
\end{align}
where $\mathbf{0}$ is an all zero tensor. 
With all the above loss functions, we get the final loss function to optimize the CNN, \ie, 
\begin{align} \label{eq:consist_loss}
\mathcal{L}  &= \lambda_1\mathcal{L}_\text{con1}(\mathbf{I}_1,\mathbf{I}_2,\tilde{\mathbf{I}}_1,\tilde{\mathbf{I}}_2) +\mathcal{L}_\text{con}(\mathbf{I}_1,\mathbf{I}_2,\mathbf{B}_1,\mathbf{B}_2),\nonumber\\
&+\mathcal{L}_\text{con}(\tilde{\mathbf{I}}_1,\tilde{\mathbf{I}}_2,\mathbf{B}_1,\mathbf{B}_2),
\end{align}
%
where $\mathcal{L}_\text{con}(\mathbf{I}_1,\mathbf{I}_2,\mathbf{B}_1,\mathbf{B}_2)$ is defined as, \ie,
%
\begin{align} \label{eq:consist_combine}
\mathcal{L}_\text{con}  = \lambda_2\mathcal{L}_\text{con2} + \lambda_3\mathcal{L}_\text{con3} +\lambda_4\mathcal{L}_\text{con4}+\lambda_5\mathcal{L}_\text{con5}.   
\end{align}
and $\{\lambda_i\}$ are to balance different terms.
The last term of \reqref{eq:consist_loss} is to utilize the $\mathcal{L}_\text{con}$ on $(\tilde{\mathbf{I}}_1,\tilde{\mathbf{I}}_2,\mathbf{B}_1,\mathbf{B}_2)$, that is, we can conduct the same operations and loss functions from \reqref{eq:bg_replace1} to \reqref{eq:consist_loss5} on the augmented image pair $<\tilde{\mathbf{I}}_1,\tilde{\mathbf{I}}_2>$ by producing  $<\tilde{\mathbf{I}}_1',\tilde{\mathbf{I}}_2'>$, and $<\tilde{\mathbf{B}}_1',\tilde{\mathbf{B}}_2'>$ via \reqref{eq:bg_replace1}, and \eqref{eq:bg_replace3}, and calculating corresponding loss functions.

\subsection{Learning Algorithm}
\label{subsec:algorithm}

\textbf{Algorithm (\ie, Line 12-20 in Algorithm \ref{alg}).} At the $t$ iteration during training, we sample an image pair from the training dataset $\mathcal{T}_\text{train}$ and perform the change detection via $\phi_\theta(\cdot)$ (See Line 14 in Algorithm \ref{alg}). Then, we generate an augmented image pair via the BGMix function. After that, we calculate the loss functions defined in \secref{subsec:augloss} (See Line 16-20 in Algorithm \ref{alg}) and update the model parameters.

\textbf{Implementation details.} 
We set the number of augmentation paths $K$ to 4 and the augmentation operation set $\mathcal{O}$ with eight kinds of operations, \ie, $\mathcal{O}$=\{background-aware operation, auto contrast, equalize, posterize, rotate, solarize, shear, translate\}. The hyper-parameters $\lambda_1$, $\lambda_2$, $\lambda_3$, $\lambda_4$ and $\lambda_5$ are set to 1, 1, 5, 1 and 0.01 for AICD, and 1, 1, 3, 1 and 0.01 for BCD, respectively.
%
%
The proposed method is implemented with Pytorch~\cite{paszke2019pytorch}. All experiments in the subsequent section are conducted on a 24GB RTX3090 GPU. We use SGD to optimize the network parameters. The batch size, learning rate, and momentum are set to 4, 1e-4, and 0.5, respectively. The largest iteration number is 100k.

\section{Experimental Results}
\label{sec:experiment}

Due to the space limitation, we report more analysis results in the supplementary material including visualization results, ablation study of BGMix, \etc.

\subsection{Setup}
\textbf{Baselines.}
We use BGMix to enhance four change detection methods, \ie, AFA~\cite{ru2022learning},  FCD~\cite{wu2022fully}, WCDNet~\cite{andermatt2020weakly}, and SG~\cite{zhao2021self}, and compare with three SOTA data augmentation methods, \ie, CutMix~\cite{yun2019cutmix},  MixUp~\cite{huang2020self} and AugMix~\cite{hendrycks2020augmix}. 

FCD and WCDNet are two SOTA weakly supervised change detectors.
AFA is a weakly supervised segmentation method that has excellent performance on semantic segmentation tasks. We modify the single-image input of AFA into double-image input to make it suitable for CD tasks.
SG is a SOTA weakly supervised defocus blur detection method via dual adversarial discriminators. We modify the architecture of SG for change detection by changing the input and output layers. This variant is named SGCD. Details can be found in the supplementary material. 

%


\textbf{Datasets.} We conduct the experiments on two widely used remote sensing CD datasets, \ie AICD~\cite{bourdis2011constrained} and BCD~\cite{ji2018fully}.
AICD~\cite{bourdis2011constrained} has 1000 aerial image pairs with the resolution of $800\times600$. 
We randomly select 900 image pairs as the training dataset and the rest for testing. For facilitating the training and inference, all images are cropped with $256\times 256$ by sliding window manner without overlap except at the image boundaries. Following \cite{chen2021adversarial}, we paste the changes cropped from changing image pair onto the background image pairs to balance the ratio of the changing and background image pairs. After pasting, we obtain 8103 and 903 training and testing image pairs, respectively.

BCD~\cite{ji2018fully} is constructed based on a high-resolution image pair with $32507\times15354$. We crop the image pairs having a resolution of $256\times 256$ with a sliding window manner from the high-resolution image pair. We randomly select 90\% from the cropped image pairs as the training dataset and the left is used as the testing set.

Note that, we follow the setups of BYOL~\cite{grill2020bootstrap} and adopt random flip and color distortion to enrich the training examples of the two training datasets (\ie, AICD and BCD). After this operation, the ratio of changing image pairs to background images in each training dataset is about to $1:1$. Finally, we obtain 24309 and 2709 training and testing image pairs for AICD, and 15570 and 1728 training and testing image pairs for BCD, respectively.
All compared data augmentation methods are used to further augment these training examples with their own augmentation strategies and loss functions.






\textbf{Evaluation Metrics.} We adopt F1-score (F1), overall accuracy (OA), and Intersection over Union (IoU) as the metrics for evaluation.

%




\begin{table}[!tb]
    \centering
    \caption{Quantitative comparison of the WSCD methods with different state-of-the-art data augmentation methods  on AICD and BCD. The best results are marked in \textbf{bold}.}
     \resizebox{\linewidth}{!}{
    \begin{tabular}{lcccccc}
    \toprule
    
    \multirow{2}{*}{Method} 
    & \multicolumn{3}{c}{AICD} & \multicolumn{3}{c}{BCD}\\
    \cline{2-4} \cline{5-7} 
     & F1& OA& IoU& F1& OA& IoU \\
    \midrule

    
    \multicolumn{3}{l}{\textbf{AFA~}\cite{ru2022learning}} & &  &  & \\
    
    w/o Aug        & 0.037 & 0.799 & 0.017 & 0.379 & 0.651 & 0.212 \\
    AugMix & 0.327 & 0.960 & 0.195 & 0.315 & 0.527 & 0.168 \\
    CutMix & 0.172 & 0.982 & 0.092 & 0.284 & 0.451 & 0.151 \\
    MixUp  & 0.375 & 0.951 & 0.229 & 0.262 & 0.349 & 0.136 \\ 
    BGMix  & \textbf{0.496} & \textbf{0.988} & \textbf{0.324} & \textbf{0.405} & \textbf{0.701} & \textbf{0.243} \\
    \midrule
    \multicolumn{4}{l}{\textbf{FCD}~\cite{wu2022fully}}  & & & \\
    w/o Aug& 0.373& \textbf{0.995} & 0.210 &0.285&\textbf{0.826}&0.150 \\
    AugMix & 0.421 & 0.989 & 0.223 & 0.269 & 0.792 & 0.142 \\
    CutMix & 0.298 & 0.993 & 0.176 & 0.305 & 0.791 & 0.171 \\
    MixUp & 0.310 & 0.827 & 0.189 & 0.357 & 0.801 & 0.211 \\
    BGMix &\textbf{0.470} &0.993 &\textbf{0.317} & \textbf{0.371} & 0.813 & \textbf{0.223} \\
    \midrule
    \multicolumn{5}{l}{\textbf{WCDNet}~\cite{andermatt2020weakly}}  &  & \\
    w/o Aug  & 0.605 & 0.993 & 0.430 & 0.393 &	0.823 & 0.221\\
    AugMix & 0.247 & 0.994 & 0.141 &0.031 & 0.810 & 0.014 \\
    CutMix & 0.579 & 0.992 & 0.407 & 0.006 & \textbf{0.869} & 0.003 \\
    MixUp  & 0.050 & 0.995 & 0.021 & 0.160 & 0.529 & 0.054 \\
    BGMix  & \textbf{0.749} & \textbf{0.996} & \textbf{0.521} & \textbf{0.453} & 0.835 & \textbf{0.276} \\
    \midrule
    
    \multicolumn{5}{l}{\textbf{SGCD}~\cite{zhao2021self}} & & \\
    w/o Aug&0.771 & 0.997 & 0.612 & 0.431 & 0.559 & 0.233 \\
    AugMix & 0.712 & 0.995 & 0.546 & 0.297 & 0.796 & 0.136 \\
    CutMix & 0.747 & 0.996 &	0.586 & 0.358 & 0.470 & 0.200\\
    MixUp& 0.576 & 0.996 & 0.440 & 0.259 & 0.174 & 0.171\\
    BGMix  &\textbf{0.854} &\textbf{0.998} & \textbf{0.736} & \textbf{0.624} & \textbf{0.844} & \textbf{0.427}\\

    
    \bottomrule
    \end{tabular}
    }
    \label{tab:AICD_BCD}
\end{table}



\subsection{Comparison of Data Augmentations}

\tableref{tab:AICD_BCD} shows the results of different WSCD methods with the state-of-the-art data augmentation methods on AICD and BCD datasets.
Based on the table, we have the following observations. \ding{182} The compared data augmentation methods have different effects for different CD methods and datasets.
Specifically, among the compared methods, MixUp achieves the highest F1 and IoU improvements for AFA and FCD on AICD and BCD, respectively. 
AugMix achieves the highest F1 and IoU improvements for FCD on AICD.
\ding{183} Not all data augmentation methods can improve the performance of the change detectors.
The F1 score of FCD with CutMix is 0.298 on AICD, which is lower than the F1 score~(0.373) without data augmentation.
On AICD and BCD, none of the state-of-the-art data augmentation methods improves the detection performance of WCDNet.
\ding{184} Our BGMix achieves the highest performance improvements for all the change detectors on both datasets. On AICD, BGMix improves F1 scores of AFA, FCD, WCDNet and SGCD to 0.496, 0.470, 0.749 and 0.854, respectively.
On BCD, BGMix improves F1 scores of AFA, FCD, WCDNet and SGCD to 0.405, 0.371, 0.453 and 0.624, respectively.
Results of \tableref{tab:AICD_BCD} manifest the effectiveness of our method.


As shown in \figref{fig:my_vision}, the changes in AICD are much smaller than the changes in BCD.
Trained with BGMix, the change detectors can effectively suppress the background changes and highlight the real changes (See the results of AFA in \figref{fig:my_vision}). 
BiGMix also can increase the recall of the detected changes for FCD~(See the 2nd-4th rows of \figref{fig:my_vision}) and WCDNet~(See the 3rd and 4th rows of \figref{fig:my_vision}).
As shown in the last two columns of \figref{fig:my_vision}, the details of the detected changes of SGCD trained with BGMix are clearer than the results of SGCD without data augmentation.

\subsection{Ablation Study of Loss Functions}
%
%
SGCD obtains better CD performances than other change detectors on both datasets. Thus, we use SGCD for the ablation study. 
We remove each term of the augmented \& real data consistency loss and retain other losses to verify the effect of each loss proposed in \secref{subsec:augloss}.
As shown in \tableref{tab:abla_sscl}, SGCD collapses on the two datasets when removing $\mathcal{L}_\text{con2}$ loss, which demonstrates that the qualities of the synthesized changing and background image pairs are critically important.
Removing losses of $\mathcal{L}_\text{con1-5}$ results in different degrees of performance degradation on the two datasets. While using the proposed loss achieves the best performances on both datasets, which manifests the usefulness of our loss.


\begin{table}[t]
    \centering
    \caption{Ablation study of augmented \& real data consistency loss with SGCD on AICD and BCD. The best results are marked in \textbf{bold}.} 
    \begin{tabular}{lcccc}
    \toprule
    Dataset&Method & F1&OA& IoU \\
    \midrule
    \multirow{5}{*}{AICD} &
    w/o  $\mathcal{L}_\text{con1}$ &0.635 & 0.992&0.449\\
    &w/o $\mathcal{L}_\text{con2}$ &0.019 & 0.994 & 0.009\\
    &w/o $\mathcal{L}_\text{con3}$ & 0.641 & 0.993 & 0.454\\
    &w/o $\mathcal{L}_\text{con4}$ &0.818 &0.998 &0.674 \\
    &w/o $\mathcal{L}_\text{con5}$ &0.849 &0.998 &0.724 \\
    & $\mathcal{L}_\text{con}$  & \textbf{0.854}& \textbf{0.998}&	\textbf{0.736} \\
    \midrule
    \multirow{5}{*}{BCD} &
    w/o  $\mathcal{L}_\text{con1}$ &0.362 & 0.807 & 0.170\\
    &w/o $\mathcal{L}_\text{con2}$ & 0.002 & 0.749 & 0.001\\
    &w/o $\mathcal{L}_\text{con3}$ & 0.311 & 0.640 & 0.152\\
    &w/o $\mathcal{L}_\text{con4}$ & 0.347 & 0.809 & 0.166\\
    &w/o $\mathcal{L}_\text{con5}$ &0.428 &0.818 &0.225 \\
    & $\mathcal{L}_\text{con}$ & \textbf{0.624} & \textbf{0.844} & \textbf{0.427} \\

    \bottomrule
    \end{tabular}
    \label{tab:abla_sscl}
    \vspace{-5px}
\end{table}

\section{Conclusion}
\label{sec:conclusion}
In this paper, we study the generalization issue of change detection from data augmentation and develop a novel weakly supervised training algorithm. The proposed background-mixed augmentation is specifically designed for change detection by augmenting examples under the guidance of a set of background images and letting targeted deep models see diverse background variations. Moreover, we propose the \textit{augmented \& real data consistency loss} to encourage the generalization increase significantly. We use the proposed method as a general framework to enhance four SOTA  change detectors.
Extensive experimental results demonstrate the superiority of our method over other SOTA  data augmentation methods on the change detection task.
In the future, we can use recent adversarial attacks \cite{guo2020watch,tian2021ava,zhai2020s} to guide the augmentation and further enhance the detection accuracy. 

\noindent\textbf{Acknowledgment:} The proposed research is supported by A*STAR Centre for Frontier AI Research. It is also supported by the National Research Foundation Singapore and DSO National Laboratories under the AI Singapore Programme (AISG Award No: AISG2-RP-2020-019), Singapore National Cybersecurity R\&D Program No. NRF2018NCR-NCR005-0001, National Satellite of Excellence in Trustworthy Software System No.~NRF2018NCR-NSOE003-0001, NRF Investigatorship No.~NRF-NRFI06-2020-0001, and in part by the Scientific Research Project of Tianjin Educational Committee
(Grant No. 2021ZD002). We gratefully acknowledge the support of NVIDIA AI Tech Center (NVAITC).


\bibliography{aaai22}
\clearpage
\section{Appendix}
\subsection{Implementation details of SGCD}
\label{sec:sgcd}
SGCD is derived from a weakly supervised defocus blur detection, \ie, SG~(Zhao, Shang, and Lu 2021). SG consists of a generator and two discriminators. Directly using SG for change detection cannot find changes on AICD and BCD datasets. 
Thus, we make a simple change of the encoder of SG to take image pairs, and then extract features $\mathbf{F}_1^i$ and $\mathbf{F}_2^i$ from $\mathbf{I}_1$ and $\mathbf{I}_2$ by the $i$th convolutional module, respectively. Intuitively, we need to calculate the difference features between $\mathbf{F}_1$ and $\mathbf{F}_2$ at corresponding level for locating change regions. The difference features $\mathbf{D}_i$ can be computed by 
\begin{equation}
    \mathbf{D}_i=\Psi(|\mathbf{F}_1^i -\mathbf{F}_2^i|), i\in(1,...,5) \nonumber
\end{equation}

where $\Psi(\cdot)$ denotes convolutional operations.
We use $\mathbf{D}_i$ to replace the original feature of the $i$th convolutional module of SG's decoder to construct a feasible change detector, \ie, SGCD. 
With the simple modifications, SGCD achieves the best change detection performances on both datasets.

Although SGCD achieves the highest F1 values, 0.771 and 0.431 on AICD and BCD than other compared change detectors, respectively. Our method, \ie, BGMix can further improve the F1 values of SGCD to 0.854 and 0.624 on AICD and BCD, respectively.

\subsection{Ablation study of Eq.~(13)} 
\label{sec:abla13}
Eq.~(13) has three sub-losses, \ie{}, $\mathcal{L}_\text{con1}(\mathbf{I}_1,\mathbf{I}_2,\tilde{\mathbf{I}}_1,\tilde{\mathbf{I}}_2)$, $\mathcal{L}_\text{con}(\mathbf{I}_1,\mathbf{I}_2,\mathbf{B}_1,\mathbf{B}_2)$ and $\mathcal{L}_\text{con}(\tilde{\mathbf{I}}_1,\tilde{\mathbf{I}}_2,\mathbf{B}_1,\mathbf{B}_2)$.  On AICD, removing $\mathcal{L}_\text{con1}(\mathbf{I}_1,\mathbf{I}_2,\tilde{\mathbf{I}}_1,\tilde{\mathbf{I}}_2)$ hurts the performance of SGCD most. While on BCD, removing $\mathcal{L}_\text{con}(\tilde{\mathbf{I}}_1,\tilde{\mathbf{I}}_2,\mathbf{B}_1,\mathbf{B}_2)$ obtains the worst change detection performance. Combining all the losses as Eq.~(13) achieves the best change detection results on both datasets.

\begin{table}[!htb]
    \centering
    \captionsetup{font={small}}
    \caption{Ablation study of Eq.~(13) with SCCD. The best results are marked in \textbf{bold}.}
    \resizebox{\linewidth}{!}{
    \begin{tabular}{ccccccc}
    \toprule
    & \multicolumn{3}{c}{AICD} & \multicolumn{3}{c}{BCD} \\
    \cline{2-4} \cline{5-7}
    & F1 & OA & IoU & F1 & OA & IoU \\
    \midrule
        w/o  $\mathcal{L}_\text{con1}(\mathbf{I}_1,\mathbf{I}_2,\tilde{\mathbf{I}}_1,\tilde{\mathbf{I}}_2)$ &0.635 & 0.992&0.449&0.362 & 0.807 & 0.170\\
        w/o $\mathcal{L}_\text{con}(\mathbf{I}_1,\mathbf{I}_2,\mathbf{B}_1,\mathbf{B}_2)$ & 0.814 & 0.998 & 0.699 & 0.430 & 0.820 & 0.223  \\
        w/o $\mathcal{L}_\text{con}(\tilde{\mathbf{I}}_1,\tilde{\mathbf{I}}_2,\mathbf{B}_1,\mathbf{B}_2)$ &0.809 & 0.997 & 0.671 & 0.210 & 0.803 & 0.092  \\
        ALL  &\textbf{0.854} &\textbf{0.998} & \textbf{0.736} & \textbf{0.624} & \textbf{0.844} & \textbf{0.427}\\
    
        \bottomrule
    \end{tabular}}

    \label{tab:loss_item}
\end{table}

In practice, we initialize ($\lambda_i, i=1,...,5$ as 1.0 and tune $\lambda_3$ and $\lambda_5$ since the corresponding loss scores are quite different from the ones of $\lambda_1$, $\lambda_2$, and $\lambda_4$. To address the concerns on hyper parameters selection, we conduct experiments by training SGCD detector with our method and changing $\lambda_1$, $\lambda_2$, $\lambda_4$ to 0.1, 10, $\lambda_3$ to 3, 10, and $\lambda_5$ to 0.001, 0.1 on AICD dataset. In the \tableref{tab:params} and \tableref{tab:abla_sscl}, our method is less sensitive to  $\lambda_1$ and $\lambda_3$ while changing $\lambda_2$, $\lambda_4$ and $\lambda_5$ leads to an obvious performance drop. 

\begin{table}[!ht]
    \centering
    \caption{Caption}
    \resizebox{0.7\linewidth}{!}{
    \begin{tabular}{ccccc}
    \toprule
$\lambda_i$	&thres&	F1&	OA&	IoU\\
\midrule
\multirow{3}{*}{$\lambda_1$}	&0.1&0.836&	0.998&	0.701\\
	&1&	0.854&	0.998&	0.736\\
	&10&0.821&	0.998&	0.672\\
	\midrule
\multirow{3}{*}{$\lambda_2$}	&0.1&0.704&	0.997&	0.578\\
	&1&	0.854&	0.998&	0.736\\
	&10&0.298&	0.959&	0.163\\
	\midrule
\multirow{3}{*}{$\lambda_3$}	&3&	0.808&	0.998&	0.666\\
	&5	&0.854&	0.998&	0.736\\
	&10	&-&	-&	-\\
	\midrule
\multirow{3}{*}{$\lambda_4$}&0.1&0.647&	0.993&0.465\\
	&1	&0.854&0.998&	0.736\\
	&10	&0.640&	0.993&	0.457\\
	\midrule
\multirow{3}{*}{$\lambda_5$}&0.001&0.703&0.995&0.530\\
	&0.01&0.854&0.998&	0.736\\
	&0.1&0.701&	0.995&	0.524\\
	\bottomrule
    \end{tabular}}
    
    \label{tab:params}
\end{table}

\subsection{Path number selection in BGMix}
\label{sec:path}
We adopt 4 paths for BGMix in our experiments. In this supplementary material, we report more results of SGCD using BGMix with different path numbers in \tableref{tab:path_num}. Note that the residual path is kept for all experiments. Thus, the path number is started from 2 and ended at 6. We can find that the results of SGCD with 2 and 3 paths on both datasets are obviously worse than the results of SGCD with other path numbers. Increasing path number to 5 and 6, SGCD can not further improve the values of different metrics. It is reasonable that small path number cannot make BGMix generate promising image pairs, while large path number breaks the distribution of the real change image pairs. Moderate path number, \ie{}, 4, is suitable for obtaining high-quality change detection results.

\begin{table}[!h]
    \centering
    \large
    \caption{The effects of the path number (\#Num) in BGMix. The best scores are marked in \textbf{bold}}
    \resizebox{\linewidth}{!}{
    \begin{tabular}{ccccccc}
    \toprule
    &\multicolumn{3}{c}{AICD} & \multicolumn{3}{c}{BCD} \\
    \cline{2-4} \cline{5-7}
    \#Num & F1 & OA & IoU & F1 & OA & IoU \\
    \midrule
     2 & 0.651 & 0.993 & 0.463 & 0.390 & 0.817 & 0.203 \\
     3 & 0.637 & 0.993 & 0.450 & 0.384 & 0.803 & 0.190 \\
     4 & 0.854 & 0.998 & 0.736 & 0.624 & 0.844 & 0.427 \\
     5 & 0.723 & 0.997 & 0.601 & 0.215 & 0.807 & 0.098 \\
     6 & 0.822 & 0.998 & 0.686 & 0.478 & 0.827 & 0.267 \\
    \bottomrule
    \end{tabular}}

    \label{tab:path_num}
\end{table}


\subsection{Ablation study of augmentation operations}
\label{sec:operation}
We test the performance of SGCD with BGMix by removing one of the operations on AICD dataset. As shown in \tableref{tab:ablation_operations}, removing the background-aware augmentation operation from operation set $\mathcal{O}$, the values of F1 and IoU drop obviously than removing other operations, which manifests the effectiveness of the proposed background-aware augmentation operation.

\begin{table}[h]
    \centering
    \captionsetup{font={small}}
     \caption{Ablation study of augmentation operations BGMix. `bg-aware' denotes the background-aware augmentation operation.}
    \begin{tabular}{lccc}
    \toprule
    Operation set &F1&OA&IoU  \\
    \midrule
     w/o translate & 0.819 &  0.998 & 0.688 \\
     w/o shear & 0.821 & 0.998 & 0.685 \\
     w/o autocontrast & 0.813 & 0.998 & 0.666 \\
     w/o equalize & 0.835 & 0.998 & 0.707 \\
     w/o posterize & 0.802 &  0.998 & 0.655 \\
     w/o rotate   & 0.846 & 0.998 & 0.721 \\
     w/o solarize & 0.841 & 0.998 & 0.714 \\
     w/o bg-aware & 0.712 & 0.995 & 0.546 \\
     \bottomrule
    \end{tabular}
   
    \label{tab:ablation_operations}
\end{table}

\subsection{More Visualization results}
\label{sec:visual}
\begin{figure*}[!ht]
    \centering
    \captionsetup{font={small}}
    \includegraphics[width=0.96\linewidth]{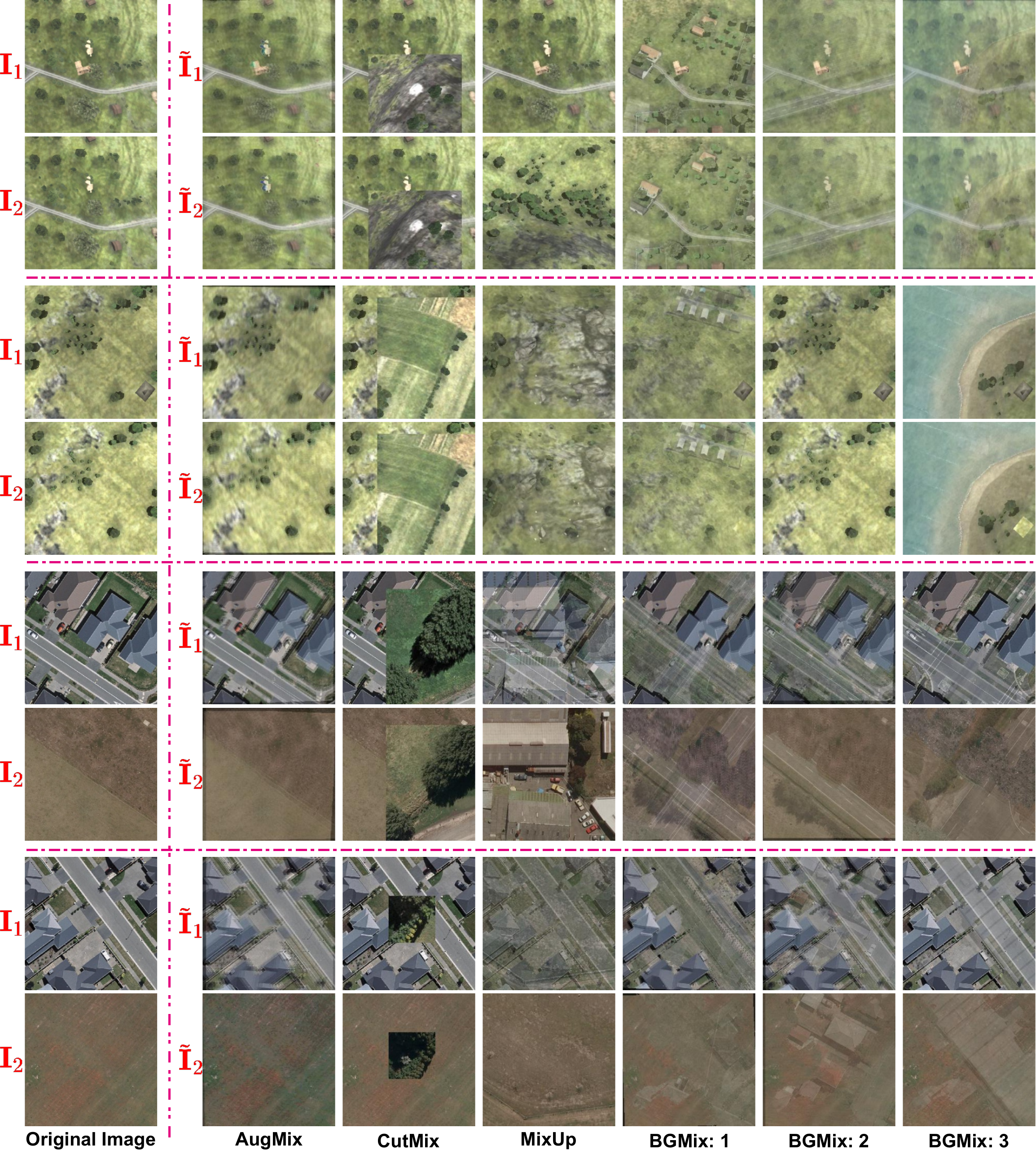}
    \caption{Examples of augmented image pairs of different data augmentation methods. We show 3 augmented pairs generated by BGMix for each input image pair.}
    \label{fig:vial_aug}
\end{figure*}

More examples of augmented image pairs of different data augmentation methods are shown in \figref{fig:vial_aug}. Augmix and MixUp transform the whole image with different degradation operations, which may hurt the representation of the real changes. CutMix cuts the image contents from other image pairs, which might occludes the changes. On the contrary, the augmented image pairs not only introduce different background contents, but also degrades the images randomly.

\begin{figure*}[!ht]
    \centering
    \captionsetup{font={small}}
    \includegraphics[width=\linewidth]{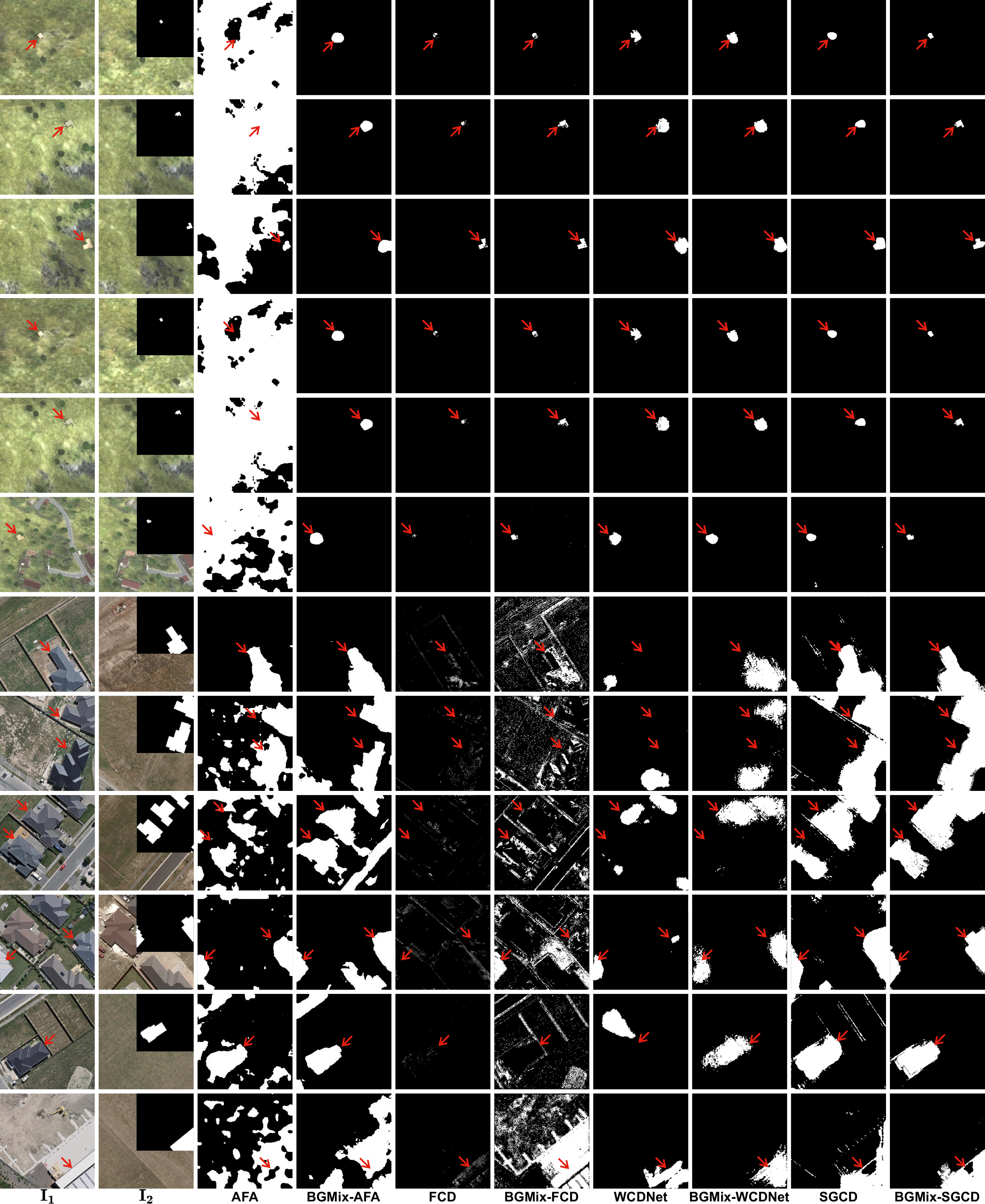}
    \caption{Visual comparison of CD results detected by different methods with and without BGMix. The change objects are marked by the red arrows.}
    \label{fig:suppl_results}
\end{figure*}

We show more results of different change detectors with and without BGMix in \figref{fig:suppl_results}. The changes of these examples are seated at the boundaries, centers, or corners of the image pairs, and have different sizes and challenge illuminations. Obviously, BGMix improves the change detectors to obtain better results.

\end{document}